\documentclass[letterpaper]{article}
\usepackage[preprint]{aaai2027}
\usepackage[hyphens]{url}
\usepackage{graphicx}
\usepackage{natbib}
\usepackage{caption}
\usepackage{amsmath}
\usepackage{algorithm}
\usepackage{algorithmic}
\usepackage{newfloat}
\usepackage{listings}
\usepackage{booktabs}
\usepackage{multirow}

\newcommand{\approach}{actuation-slack refresh}
\newcommand{\Approach}{Actuation-slack refresh}

\title{The Gate, Not the Cache: Gate Provenance Bounds the Closed-Loop Reliability of Training-Free VLA Token Skipping}
\author{
    Qi Luo\textsuperscript{\rm 1},
    Shuaijun Liu\textsuperscript{\rm 1},
    Hao Zhao\textsuperscript{\rm 1},
    Kunlin Li\textsuperscript{\rm 1},
    Xiaobo Wang\textsuperscript{\rm 2,3},
    Ningxing Su\textsuperscript{\rm 1},
    Dongsheng Wang\textsuperscript{\rm 4},
    Yun Chen\textsuperscript{\rm 1}\corresponding
}
\affiliations{
    \textsuperscript{\rm 1}The Hong Kong University of Science and Technology (Guangzhou)\\
    \textsuperscript{\rm 2}Shenzhen University of Advanced Technology\\
    \textsuperscript{\rm 3}Sangfor Technologies Inc.\\
    \textsuperscript{\rm 4}Tsinghua University\\
    \{qluo615, sliu529, hzhao688, kli082\}@connect.hkust-gz.edu.cn,
    ningxinsu@hkust-gz.edu.cn,\\
    wds@tsinghua.edu.cn, wangxiaobo@suat-sz.edu.cn,
    yunchen@hkust-gz.edu.cn
}

\begin{document}

\maketitle

\begin{abstract} 
Token skipping is a widely used training-free way to accelerate vision--language--action (VLA) models by bypassing computation for most visual tokens at each control step according to a gate. When the next gate is harvested from the previous accelerated forward, however, the tokens skipped at one step are also the ones least visible to the next gate, and the damage can compound across control steps until the task fails. We study the two mechanisms this class is built on, reuse and deletion, crossing each against where its gate signal comes from on identical episodes. At a skip ratio of 0.9 on LIBERO-Object, both collapse when the gate comes from the model's own accelerated forwards, to 0.68 under reuse and to 0.31 under deletion against a dense 1.00, and the collapse is invisible to the action-level detectors we evaluate. What separates collapse from dense-level operation is not the mechanism but whether the gate is clean, computed by a forward that skipped nothing. We therefore propose \textbf{\approach}, one dense pass run while the robot executes its current action chunk, off the critical path, that hands the next step a clean gate and a fresh KV base. Since the measured detectors do not reliably reveal the failure, the refresh is unconditional rather than triggered. Both mechanisms then recover to 0.98, keeping the speed of skipping and the information of a dense pass. We then integrate the refresh into state-of-the-art caching and pruning methods across two VLA policies, 4 LIBERO suites, and 4 SIMPLER tasks, where it repairs every collapse caused by using a self-harvested gate. Serve latency drops 18--22\% below dense, measured both in simulation and on a physical robot. Where the gate signal comes from, not how tokens are skipped, decides closed-loop reliability for accelerated VLAs.
\end{abstract}

\section{Introduction}
\label{sec:intro}

Vision--language--action (VLA) models \citep{black2025pi0,geminiroboticsteam2025gemini,nvidia2025groot,kim2025finetuning,li2024cogact} sit at the frontier of embodied intelligence, letting a robot carry out a natural-language instruction through continuous manipulation. A typical policy turns the current camera view into a few hundred visual tokens, runs one forward pass over them together with the instruction, and emits a short action chunk. The robot executes it, the view changes, and the policy runs again. Every pass through this loop pays the full inference cost of a large vision--language backbone, much of it spent on visual tokens that barely change between frames. Training-free token skipping promises to reclaim that cost. A gate marks which tokens to skip, and the marked ones are either served from a cross-step cache, reusing the KV tensors computed for them at the previous step \citep{xu2025vlacache}, or deleted outright, on the grounds that patches like static background carry little task information \citep{wang2026specprune,liu2025bridging}. Published methods report near-lossless success \citep{cheng2026vlaiap,chen2026fast}.

\begin{figure}[t]
\centering
\includegraphics[width=0.9\columnwidth]{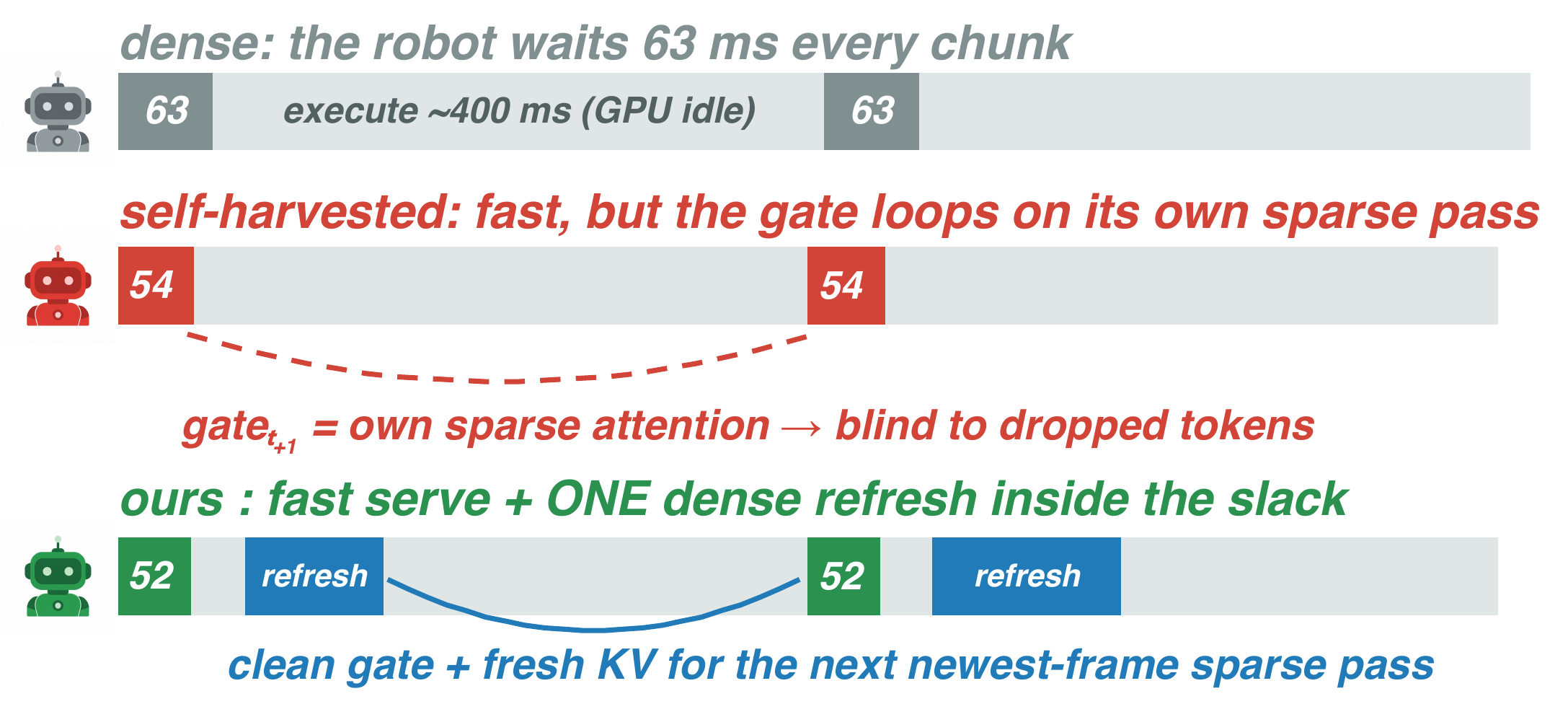}
\caption{One control chunk under dense serving, self-harvested skipping, and \approach{} (ours).}
\label{fig:teaser}
\end{figure}

As shown in Fig.~\ref{fig:teaser}, \emph{dense} serving skips nothing and spends ${\sim}63$\,ms on the forward (the 7B OpenVLA-OFT policy on LIBERO, H100), after which the GPU sits idle for the ${\sim}400$\,ms the robot needs to execute the chunk. Deployed skipping computes only the tokens the gate keeps and cuts the serve to ${\sim}54$\,ms. When the gate for the next chunk is read from that same sparse forward, its signal comes from a forward in which many tokens were already stale or absent. We call such a gate self-harvested.

We run both skipping mechanisms, reuse and deletion, twice over identical episodes, once with the gate taken from the accelerated forward that served the current action chunk and once with it taken from a dense pass whose action output was discarded. At a skip ratio of 0.9 on LIBERO-Object \citep{liu2023libero}, the self-harvested gate takes paired success from a dense 1.00 down to 0.68 under reuse and to 0.31 under deletion, while the gate from the dense pass holds both at dense level. The failure is silent. Single-step action deviation stays negligible, and every reactive detector we instrumented scores near chance. Concurrent detectors built for distribution shift \citep{pan2026vlacorrector,seligmann2026vlafail,mahato2026early} remain untested here. The failure-critical shift is concentrated in tokens that the sparse pass no longer recomputes. Methods that look lossless at their published operating points can cross a failure cliff that their own signals do not reveal.

Prior work optimizes gate selection and the choice between cache reuse and deletion, but to our knowledge no prior work isolates gate provenance from the skipping mechanism in closed loop. The loop that blinds the gate also holds the fix. We spend that idle window on one dense pass over the same frame, which serves no action and therefore hands the next chunk a gate computed over all tokens and a KV base that no sparse forward has touched (Fig.~\ref{fig:teaser}, bottom). It runs on every chunk and off the critical path, so it costs no serve latency. Added to VLA-Cache \citep{xu2025vlacache} and VLA-Pruner \citep{liu2025bridging}, the state-of-the-art methods of the caching and pruning families, it repairs every collapse caused by using a self-harvested gate across two VLA policies, 4 LIBERO suites, and 4 SIMPLER tasks while serving 18--22\% faster than dense on OpenVLA-OFT.

Refreshing on every chunk may look wasteful next to refreshing only when a trigger says so. SAFE-Pruner \citep{ma2026safepruner}, the closest method to ours, takes the second route, supplying clean saliency but deciding when to refresh based on a self-harvested consistency score. Such a trigger has to see the very shift it is meant to catch, and this one does not, firing 3 times in 3{,}222 encodes on Object while success falls to 0.19 against a dense 1.00 (Sec.~\ref{sec:safe}). The unconditional schedule makes the repair reliable, and the actuation slack keeps the refresh off the critical path.

In summary, we make the following contributions.
\begin{itemize}
\item \textbf{Isolation.} A controlled design crossing gate provenance with skipping mechanism on identical episodes shows that where the gate signal comes from, not the mechanism, separates reliable from unreliable skipping.
\item \textbf{Repair.} \Approach{}, one dense pass per chunk inside the execution window, lifts both mechanisms on Object at a 0.9 skip ratio from 0.68 and 0.31 back to 0.98, repairs every collapse caused by using a self-harvested gate once added to the named baselines, and serves 18--22\% faster than dense on OpenVLA-OFT, both in simulation and on a physical robot.
\item \textbf{Explanation.} Beyond the controlled cells, the same account grades the released methods by how much clean signal each of them still computes, which is why the shallow recompute shipped with VLA-Cache closes only part of the gap and why VLA-Pruner avoids the provenance failure until the substrate, rather than the gate, reaches its capacity boundary.
\end{itemize}

\section{Related Work}
\label{sec:related}

\paragraph{Training-free token skipping for VLAs} Two routes dominate this line. The caching route keeps every token in the attention context but serves the KV of the skipped ones from the previous step, so a patch the gate calls static still contributes to attention without being recomputed \citep{xu2025vlacache}. The pruning route removes those tokens altogether, with a saliency score deciding which patches carry too little task-relevant information to keep \citep{wang2026specprune,liu2025bridging,cheng2026vlaiap}, and token merging is a close relative that folds redundant patches together rather than dropping them \citep{li2026depthcache,chen2026fast}. Both routes optimize the spatial selection and report near-lossless success at their published operating points. In the caching and pruning systems we audit, at least one temporal component of the gate is harvested from served sparse forwards, but prior work does not isolate the resulting feedback.

\paragraph{Dense re-anchoring and triggered refresh} TTF-VLA \citep{liu2026ttfvla} interleaves periodic full-token passes on the critical path. SAFE-Pruner \citep{ma2026safepruner} caches clean saliency at key timesteps and refreshes when a consistency score $\kappa_t$ drops. In our evaluation, however, the self-harvested trigger remains largely inactive when provenance-induced failure occurs (Sec.~\ref{sec:safe}). Related work studies asynchronous, streaming, or interleaved VLA inference and real-time action correction \citep{black2025realtime,tang2025vlash,agouzoul2026understanding,sendai2025leave,guo2026reflex,huang2026environment}, with conceptual parallels to classical imprecise-computation scheduling and advanced-step control \citep{liu1991imprecise,zavala2009advanced}, but does not address gate provenance.

\paragraph{Staleness beyond robotics} Video transformers exploit temporal redundancy by selectively reprocessing changed tokens across frames \citep{dutson2023eventful}, while efficient VLM serving reuses visual KV under open-loop load \citep{qin2025vlcache}. In text-only decoding, KV capacity is managed through eviction or query-aware selection \citep{zhang2023h2o,tang2024quest}. Periodic dense rectification bounds the error a sparse-attention cache accumulates \citep{sun2025resa}, and cache-mediated feedback can amplify repetitions the model itself produced until a targeted KV intervention breaks the loop \citep{xu2026loopguard}. These settings lack the property that makes VLA skipping dangerous, an action produced by a sparse forward that moves the camera and closes the loop from what is skipped now to what is observed next.

\section{Diagnosing Silent Failure in Closed-Loop Token Skipping}
\label{sec:failure}

\subsection{Why Skipping Errors Accumulate}

A VLA policy operates in closed loop. Given an instruction such as ``pick up the black bowl and place it on the plate,'' it carries out the task over tens of chunk-level control decisions. At each such decision, the policy encodes the current camera frame into hundreds of visual tokens, runs one forward pass through the language model, and emits an action chunk. The robot executes that chunk, changing what the next step sees. The training-free methods studied here reduce language-model computation by skipping selected visual tokens, either serving cached state for them or omitting them. The fraction skipped is the \emph{skip ratio}, so at a skip ratio of 0.9 only one visual token in 10 is computed. In a closed loop, skipping also changes a future input. It perturbs the forward, the resulting action changes the next camera frame, and attention from the perturbed computation can select the next tokens to skip. Error can therefore feed back through both the environment and the selector, compound across chunks \citep{ross2011reduction}, and end in task failure.

\subsection{Gate Provenance and Skipping Mechanism}

Token skipping typically uses a \emph{gate}, a per-patch signal that decides which visual tokens are skipped at each control step. In closed loop, an independent design variable is the temporal provenance of that signal, meaning which forward supplies it. The gate for chunk $t{+}1$ can come from the sparse forward that served chunk $t$'s actions, after skipping has already withheld fresh information. We call this a \emph{self-harvested} gate. In deployment, this pattern repeats across the episode. Each served sparse forward supplies the gate and, under reuse, the KV for the next chunk. We call this the \emph{self-harvest chain}. Alternatively, the gate can come from a dense pass on the same frame whose action output is discarded. We call this a \emph{clean} gate. These provenance labels identify the signal path rather than a saliency rule.

The second variable is the skipping mechanism. Under \emph{reuse}, cached per-layer keys and values are used for skipped tokens, as in VLA-Cache \citep{xu2025vlacache}. Under \emph{deletion}, the gate retains the most salient tokens and prunes the rest, as in VLA-Pruner \citep{liu2025bridging}.

\subsection{Provenance versus Mechanism}

A deployed reuse chain mixes recursive KV reuse with a self-harvested gate. We isolate the two factors in a four-cell factorial evaluated from the same 91 initial states. Within each mechanism, we hold the saliency rule and skip ratio fixed and change only the forward that supplies the gate signal. Both reuse cells use a one-chunk-old KV base from a dense pass, removing recursive KV accumulation. The deletion cells use the same two gate provenances without KV reuse.

\begin{figure}[ht]
\centering
\includegraphics[width=0.53\columnwidth]{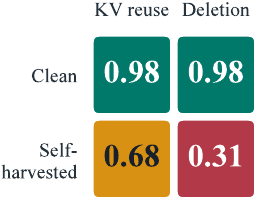}
\caption{Controlled factorial on LIBERO-Object at a 0.9 skip ratio.}
\label{fig:factorial}
\end{figure}

Fig.~\ref{fig:factorial} shows the isolated result. With a clean gate, reuse and deletion both reach 0.98 success and are statistically indistinguishable from paired dense at 1.00. Replacing only the gate source with attention harvested from served sparse forwards reduces success to 0.68 under reuse and 0.31 under deletion. Gate provenance therefore separates the reliable and collapsing rows in this controlled condition. The mechanism does not change aggregate success under the clean gate at this operating point, but it changes the severity of failure after the gate becomes self-harvested. It does not imply that every task is provenance-sensitive or that mechanism never matters. Sec.~\ref{sec:results} evaluates how far this result extends beyond the isolated condition.

Both reuse cells above hold the KV base clean by refreshing it from a dense pass every chunk, so the reuse comparison isolates gate provenance from recursive KV accumulation. The deployed reuse chain does not have this clean base because its KV base accumulates across served sparse forwards, which also supply the next gate. We therefore ask the reverse question, holding the KV base to this worst case and varying only the gate.

\begin{table}[ht]
\centering
\small
\setlength{\tabcolsep}{4pt}
\begin{tabular}{@{}lcc@{}}
\toprule
 & \textbf{KV fresh} & \textbf{KV accumulated} \\
\midrule
\textbf{self-harvested gate} & 0.68 & 0.43 \\
\textbf{clean gate} & \textbf{0.98} & \textbf{0.96} \\
\bottomrule
\end{tabular}
\caption{Gate provenance $\times$ KV provenance under reuse on LIBERO-Object at a 0.9 skip ratio, using the pure-attention gate family.}
\label{tab:gatekv}
\end{table}

Table~\ref{tab:gatekv} lets the KV base accumulate without refresh, matching the deployed chain, while the gate is either clean or self-harvested. A clean gate alone raises success from 0.43 to 0.96 on the same episodes while holding the accumulated KV base fixed. Refreshing only the KV while the gate stays self-harvested raises success to 0.68. The contrast shows that gate provenance accounts for most of the recovery, while a fresh KV base alone recovers far less.

\subsection{How Self-Harvested Gates Can Fail Silently}

\begin{figure*}[t]
\centering
\includegraphics[width=0.9\textwidth]{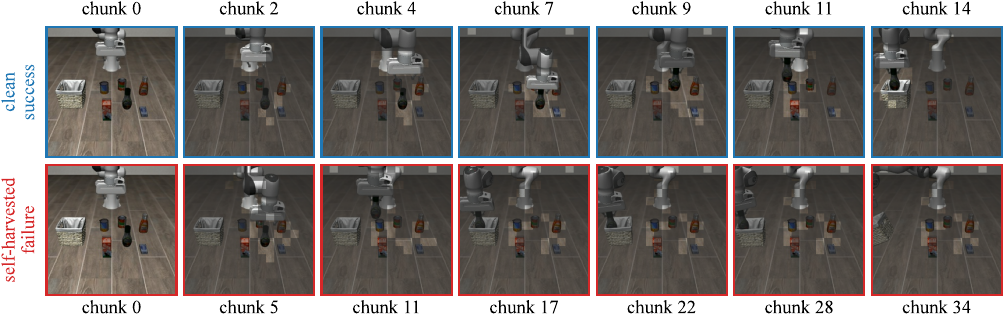}
\caption{Clean gate (top) and self-harvested gate (bottom) under deletion at a 0.9 skip ratio. Bright patches are recomputed, while dimmed patches are deleted.}
\label{fig:case}
\end{figure*}

When the next gate is self-harvested, it receives no up-to-date signal from a skipped visual token. The gate may therefore continue skipping the token even after it becomes important. Under reuse, the skipped token is represented by stale keys and values. Under deletion, the token is absent from the sparse forward and receives no attention. A self-harvested gate can therefore keep a changed but previously skipped region at a low score. The sparse pass can then carry these blind spots into the next gate, creating a second feedback loop.

Fig.~\ref{fig:case} shows this loop in a LIBERO-Object task that asks the robot to pick up the salad dressing and place it in the basket. Both rollouts begin from the same state, use deletion at a 0.9 skip ratio, and run the first chunk densely. With the clean gate, the model continues recomputing patches covering the robot arm and target object and completes the task by chunk 14. With the self-harvested gate, the selected patches drift away from task-relevant regions. The resulting blind spots persist, and the rollout reaches the 35-chunk limit without success.

The measured action-level signals do not provide useful early warning. On closed-loop rollouts reproducing the collapse, mid-chunk re-query disagreement and proprioceptive tracking deviation yield early-window AUROC of 0.45--0.65. An apparent full-episode AUROC of 0.80 is an episode-length artifact. After conditioning on duration, it becomes 0.26--0.44 because failed episodes run to the 35-chunk limit while successful episodes end at a median of 17--19. These probes do not establish that no detector can work. Concurrent feature-probe detectors target distribution shift or open-loop deviation \citep{pan2026vlacorrector,seligmann2026vlafail,mahato2026early} and remain untested here. The reactive signals we measured cannot serve as reliable refresh triggers.

These findings motivate a constructive repair. The next gate should use a signal from a dense pass over the complete token set, rather than from the served sparse path. Because the measured reactive signals provide no reliable trigger, we produce this clean signal unconditionally.

\section{Actuation-Slack Refresh}
\label{sec:method}

\begin{figure*}[t]
\centering
\includegraphics[width=0.9\textwidth]{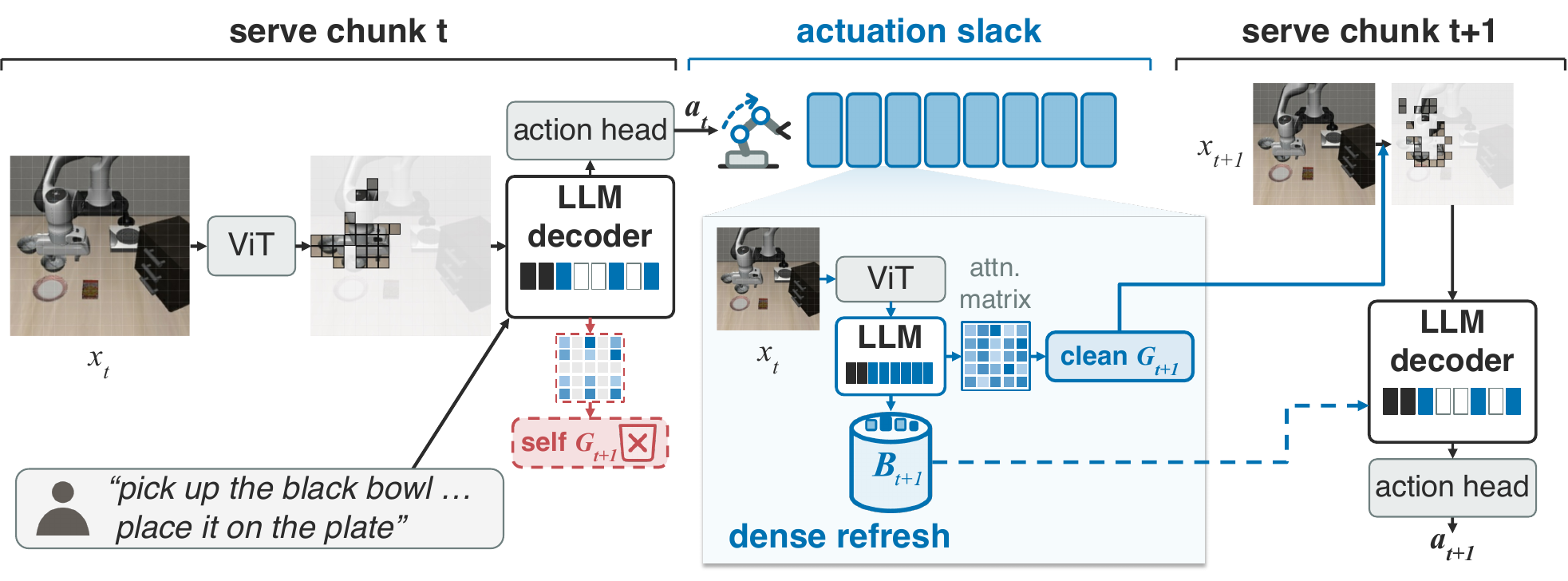}
\caption{The architecture of \approach. A dense pass overlaps action execution and supplies the next sparse serve with a clean gate signal and, for reuse, a fresh KV base.}
\label{fig:arch}
\end{figure*}

Chunked VLA policies typically alternate between one policy forward and execution of the resulting action chunk. In our setup, each chunk contains $K{=}8$ actions, whose execution takes ${\sim}400$\,ms and would otherwise leave the policy GPU idle. The policy forward lies on the serving \emph{critical path} because the robot waits for it before acting. We call the idle execution interval \emph{actuation slack}.

Two objects carry state from one chunk to the next. The gate $G_t$ marks which visual tokens are skipped at chunk $t$. The KV base $B_t$ stores per-layer keys and values for the skippable visual tokens. At chunk $t$, the critical path computes the retained visual tokens from the newest frame. Under reuse, it splices the keys and values of skipped tokens from $B_t$ into each layer. Under deletion, it removes the skipped tokens and does not use $B_t$.

Fig.~\ref{fig:arch} summarizes the workflow of \approach{}. While the robot executes the actions of chunk $t$, we run one dense pass on the same frame. Its action output is discarded, and the self-harvested temporal signal from the served sparse forward is not carried forward. The dense pass supplies the clean temporal input to the next gate and, under reuse, the next KV base,
\begin{align}
G_{t+1} &= g_r\big(A_t^{\mathrm{dense}},\, S_{t+1}^{\mathrm{live}}\big), \label{eq:gate}\\
B_{t+1} &= \big\{K_{t,\mathcal V}^{(\ell),\mathrm{dense}},\,
V_{t,\mathcal V}^{(\ell),\mathrm{dense}}\big\}_{\ell=1}^{L}, \label{eq:base}
\end{align}
Here, $\mathcal V$ is the skippable visual-token span, $A_t^{\mathrm{dense}}$ is dense full-token attention, and $S_{t+1}^{\mathrm{live}}$ is any live signal used by the target gate. The last term is omitted for the pure-attention instrument. We fix the selection rule $g_r$, including ratio $r$, and vary only its temporal attention source.

The full-token pass produces a clean temporal gate signal and, under reuse, a fresh KV base without serving an action. At chunk $t{+}1$, the sparse serve uses these one-chunk-old products while computing selected tokens from the newest frame. The first chunk runs dense because no gate or KV base is available.

A triggered refresh requires a signal that can detect changes in skipped tokens. A self-harvested trigger observes only the sparse path and can miss those changes, as Sec.~\ref{sec:failure} shows. We therefore run the dense refresh after every chunk instead of waiting for a trigger. Sec.~\ref{sec:safe} compares this design with SAFE-Pruner \citep{ma2026safepruner} and oracle-driven triggers.

\section{Experiments}
\label{sec:results}

\subsection{Setup}
\label{sec:setup}

\paragraph{Base Models} We use publicly released checkpoints for OpenVLA-OFT (OFT) \citep{kim2025finetuning}, one for each of the 4 LIBERO suites, and CogACT-Base \citep{li2024cogact}.

\paragraph{Benchmarks} We evaluate on two complementary benchmarks. LIBERO \citep{liu2023libero} is a manipulation benchmark comprising the Spatial, Object, Goal, and Long suites. On SIMPLER \citep{li2024evaluating}, we evaluate the Google Robot tasks pick-coke, move-near, drawer, and put-in-drawer. Only tokens from the primary camera view are skipped. Each comparison uses the same skip ratio across methods.

\paragraph{Metrics} Our primary metrics are success rate and inference latency. Success rate is the fraction of successful episodes. Inference latency is the critical-path serve time per policy forward.

\paragraph{Baselines} We use VLA-Cache (VC) \citep{xu2025vlacache} as the reuse baseline, VLA-Pruner (VP) \citep{liu2025bridging} as the deletion baseline, and SAFE-Pruner \citep{ma2026safepruner} as the triggered-refresh baseline. Implementation details for all 3 baselines are provided in the technical appendix. VC and VP are each evaluated without and with our refresh (+refresh). In +refresh, the slack dense pass supplies the clean gate signal to both baselines and a fresh KV base to VC.

\subsection{Closed-Loop Success}

\begin{table*}[t]
\centering
\small
\setlength{\tabcolsep}{4.5pt}
\begin{tabular}{@{}clclccccc@{}}
\toprule
\multirow{2}{*}{\textbf{substrate}} & \multirow{2}{*}{\textbf{task}} & \multirow{2}{*}{\textbf{dense}} & \multirow{2}{*}{\textbf{method}} & \multicolumn{5}{c}{\textbf{skip ratio}} \\
\cmidrule(l){5-9}
 & & & & \textbf{0.5} & \textbf{0.6} & \textbf{0.7} & \textbf{0.8} & \textbf{0.9} \\
\midrule
\multirow{10}{*}{\shortstack[c]{LIBERO $\times$\\ OpenVLA-OFT}}
 & \multirow{2}{*}{Spatial} & \multirow{2}{*}{0.93} & VC / +R & 0.93 / 0.96 & 0.94 / 0.95 & 0.93 / 0.95 & 0.91 / 0.92 & 0.91 / 0.93 \\
 & & & VP / +R & 0.80 / 0.94 & 0.92 / 0.94 & 0.88 / 0.84 & 0.90 / 0.94 & 0.90 / 0.94 \\
\cmidrule(l){2-9}
 & \multirow{2}{*}{Goal} & \multirow{2}{*}{0.96} & VC / +R & 0.93 / 0.97 & 0.93 / 0.95 & 0.96 / 0.93 & 0.96 / 0.95 & 0.98 / 0.95 \\
 & & & VP / +R & 0.99 / 0.99 & 0.99 / 0.99 & 0.99 / 0.99 & 0.96 / 0.97 & 0.94 / 0.97 \\
\cmidrule(l){2-9}
 & \multirow{2}{*}{Long} & \multirow{2}{*}{0.92} & VC / +R & 0.94 / 0.98 & 0.95 / 0.96 & 0.86 / 0.95 & 0.83 / 0.87 & 0.62 / 0.73 \\
 & & & VP / +R & 0.96 / 0.98 & 0.98 / 0.96 & 0.94 / 0.99 & 0.98 / 0.96 & 0.96 / 0.97 \\
\cmidrule(l){2-9}
 & \multirow{2}{*}{Object} & \multirow{2}{*}{1.00} & VC / +R & 1.00 / 1.00 & 1.00 / 1.00 & 1.00 / 0.99 & 0.98 / 0.99 & 0.66 / 0.96 \\
 & & & VP / +R & 1.00 / 1.00 & 1.00 / 1.00 & 1.00 / 1.00 & 1.00 / 1.00 & 0.98 / 0.98 \\
\cmidrule(l){2-9}
 & \multirow{2}{*}{Average} & \multirow{2}{*}{0.95} & VC / +R & 0.95 / \textbf{0.98} & 0.96 / \textbf{0.97} & 0.94 / \textbf{0.95} & 0.92 / \textbf{0.93} & 0.80 / \textbf{0.90} \\
 & & & VP / +R & 0.94 / \textbf{0.97} & 0.97 / 0.97 & 0.95 / 0.95 & 0.96 / \textbf{0.97} & 0.95 / \textbf{0.97} \\
\midrule
\multirow{10}{*}{\shortstack[c]{SIMPLER $\times$\\ CogACT}}
 & \multirow{2}{*}{pick-coke} & \multirow{2}{*}{0.84} & VC / +R & 0.82 / 0.92 & 0.82 / 0.85 & 0.65 / 0.74 & 0.51 / 0.61 & 0.29 / 0.37 \\
 & & & VP / +R & 0.89 / 0.89 & 0.91 / 0.89 & 0.91 / 0.88 & 0.90 / 0.88 & 0.87 / 0.84 \\
\cmidrule(l){2-9}
 & \multirow{2}{*}{move-near} & \multirow{2}{*}{0.77} & VC / +R & 0.65 / 0.62 & 0.50 / 0.55 & 0.32 / 0.43 & 0.38 / 0.43 & 0.25 / 0.29 \\
 & & & VP / +R & 0.72 / 0.73 & 0.70 / 0.77 & 0.77 / 0.70 & 0.72 / 0.70 & 0.72 / 0.68 \\
\cmidrule(l){2-9}
 & \multirow{2}{*}{drawer} & \multirow{2}{*}{0.80} & VC / +R & 0.74 / 0.77 & 0.75 / 0.75 & 0.69 / 0.80 & 0.61 / 0.76 & 0.52 / 0.69 \\
 & & & VP / +R & 0.74 / 0.75 & 0.75 / 0.75 & 0.76 / 0.78 & 0.72 / 0.74 & 0.63 / 0.60 \\
\cmidrule(l){2-9}
 & \multirow{2}{*}{put-in-drawer} & \multirow{2}{*}{0.32} & VC / +R & 0.21 / 0.21 & 0.19 / 0.23 & 0.14 / 0.16 & 0.13 / 0.14 & 0.02 / 0.02 \\
 & & & VP / +R & 0.31 / 0.39 & 0.28 / 0.37 & 0.26 / 0.20 & 0.29 / 0.23 & 0.16 / 0.19 \\
\cmidrule(l){2-9}
 & \multirow{2}{*}{Average} & \multirow{2}{*}{0.68} & VC / +R & 0.61 / \textbf{0.62} & 0.57 / \textbf{0.60} & 0.45 / \textbf{0.53} & 0.41 / \textbf{0.49} & 0.28 / \textbf{0.34} \\
 & & & VP / +R & 0.66 / \textbf{0.69} & 0.66 / \textbf{0.69} & \textbf{0.67} / 0.64 & \textbf{0.66} / 0.64 & \textbf{0.59} / 0.58 \\
\bottomrule
\end{tabular}
\caption{Closed-loop success rates across skip ratios. VC and VP denote VLA-Cache and VLA-Pruner, and +R denotes our refresh. Each cell reports baseline / +R. Dense is the full-token reference. Boldface marks the higher average.}
\label{tab:main_grid}
\end{table*}

Table~\ref{tab:main_grid} evaluates our refresh with VLA-Cache and VLA-Pruner across 4 LIBERO suites, 4 SIMPLER Google Robot tasks, and skip ratios from 0.5 to 0.9. For each benchmark, Average is the arithmetic mean of the 4 rows above it.

The largest gains appear for VC at a 0.9 skip ratio. On LIBERO, +R raises the average from 0.80 to 0.90, including Object from 0.66 to 0.96, though Long improves only from 0.62 to 0.73 and stays below the dense rate of 0.92. On SIMPLER, drawer rises from 0.52 to 0.69, and +R improves the VC average at every tested ratio. In contrast, VP shows no consistent gain from +R. On CogACT, both VP configurations remain below dense on drawer and put-in-drawer at 0.9. These results show that refresh helps when the self-harvested VC path degrades, but does not remove every capacity or gate-family boundary.

We attribute a degradation to the self-harvested gate when the self-harvested configuration degrades but its clean counterpart at the same point does not. When the clean configuration is also below dense, we report a capacity or gate-family boundary instead. Spatial VP is evaluated on a deterministic-crash survivor subset. Every collapse cell survives worst-case treatment of simulator-crash attrition. Absolute rates on crash-survivor subsets are excluded from the primary conclusions.

\subsection{Shallow versus Full Refresh}
\label{sec:repair}

\begin{table}[ht]
\centering
\small
\setlength{\tabcolsep}{3.5pt}
\begin{tabular}{@{}lcc@{}}
\toprule
\textbf{configuration} & \textbf{Object@0.9} & \textbf{drawer@0.9} \\
\midrule
dense & 1.00 & 0.80 \\
\midrule
no refresh & 0.66 & 0.52 \\
shallow recompute & 0.81 & 0.58 \\
\approach & \textbf{0.96} & \textbf{0.69} \\
\bottomrule
\end{tabular}
\caption{Refresh configurations on VLA-Cache.}
\label{tab:clean_signal}
\end{table}

Table~\ref{tab:clean_signal} compares 3 VLA-Cache configurations at a 0.9 skip ratio, with dense as the reference. OFT has 32 transformer layers. Following the released VLA-Cache design, the shallow-recompute control recomputes all visual tokens in layers 0--1 at every policy forward and reuses cached keys and values for skipped tokens in layers 2--31. The no-refresh configuration obtains 0.66 on Object and 0.52 on drawer. The shallow-recompute control reaches 0.81 and 0.58, while \approach{} reaches 0.96 and 0.69. Thus, shallow recomputation helps but does not match the full refresh.

\subsection{High-Ratio Boundary}
\label{sec:frontier}

\begin{figure}[ht]
\centering
\includegraphics[width=0.9\columnwidth]{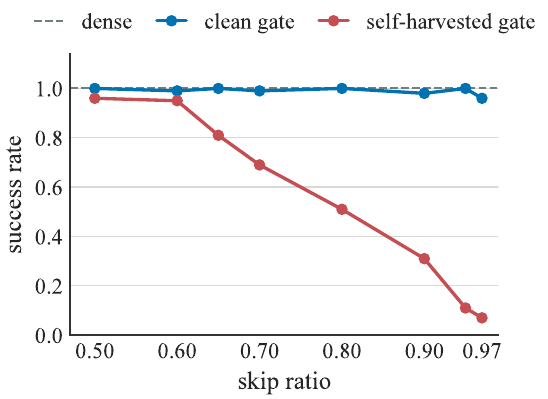}
\caption{Success rate versus skip ratio on LIBERO-Object.}
\label{fig:cliff}
\end{figure}

The impact of a self-harvested gate depends on the skip ratio and task. With deletion fixed and the pure-attention gate family used throughout, Fig.~\ref{fig:cliff} compares the two gate provenances on Object against paired dense at 1.00. The self-harvested gate is not significantly below dense through a 0.6 skip ratio, first separates at 0.65, and degrades further as the ratio rises. The clean gate remains statistically indistinguishable from dense at every tested point through 0.97. Spatial shows no significant provenance gap through 0.97. The precise boundary depends on the task, substrate, and gate family.

\begin{figure}[ht]
\centering
\includegraphics[width=0.9\columnwidth]{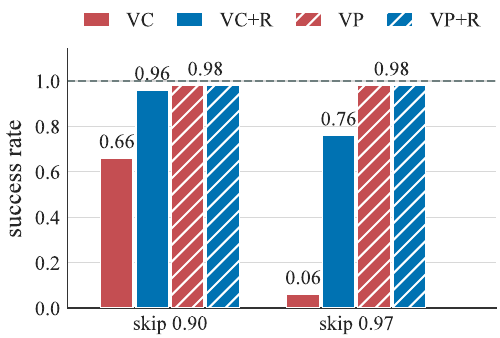}
\caption{LIBERO-Object success at skip ratios of 0.9 and 0.97. The dashed line denotes dense performance.}
\label{fig:highratio}
\end{figure}

Figure~\ref{fig:highratio} compares VC and VP on Object at skip ratios of 0.9 and 0.97. At 0.97, VC falls to 0.06, +R retains 0.76, and both VP configurations remain at 0.98. VLA-Pruner retains live gate inputs at these Object stress points. Layers 0--3 run on the full token set every step, the union selection lets the self-harvested prior add tokens but never veto them, and the diversity filter uses clean embeddings.

\subsection{Why Refresh Every Chunk}
\label{sec:safe}

We first evaluate less frequent fixed schedules. On Object, an inline dense anchor every two chunks keeps the gate at most one chunk old and achieves 1.00 success, whereas extending the interval to 4 chunks yields 0.90. A reactive trigger could avoid unnecessary anchors, but only if it detects when a refresh is needed. We test this alternative with SAFE-Pruner. It uses a dense pass to supply clean pruning saliency and drives refresh with a self-harvested trigger $\kappa_t$, the cosine between forecast and observed saliency on the surviving tokens. At a target skip ratio of 0.9, SAFE-Pruner reaches 0.95 on Spatial and 0.93 on Goal but falls to 0.66 on Long and 0.19 on Object. Table~\ref{tab:safe} shows that our clean-deletion refresh remains between 0.94 and 0.98 across the 4 suites. The trigger fires 3 times in 3{,}222 Object encodes.

\begin{table}[ht]
\centering
\small
\begin{tabular}{lcccc}
\toprule
\textbf{task} & \textbf{dense} & \boldmath$\kappa_t$ \textbf{fires} & \textbf{SAFE} & \textbf{ours} \\
\midrule
Spatial & 0.93 & 6.4\% & 0.95 & 0.94 \\
Goal & 0.96 & 5.7\% & 0.93 & 0.97 \\
Long & 0.92 & 5.4\% & 0.66 & 0.95 \\
Object & 1.00 & 0.09\% & 0.19 & 0.98 \\
\bottomrule
\end{tabular}
\caption{SAFE-Pruner and our clean-deletion refresh at a matched 0.9 skip ratio.}
\label{tab:safe}
\end{table}

On Object, the published threshold alone does not isolate provenance because an oracle trigger also rarely fires at $\gamma{=}0.92$. A threshold sweep exposes the difference. The self-harvested consistency stays at 0.98--0.99 while the clean consistency dips at failure-critical moments, because the decisive shift is concentrated in pruned tokens that $\kappa_t$ cannot see. No measured threshold achieves both dense-level success and sparse-serve latency. At $\gamma{=}0.98$, the trigger fires on 29.5\% of encodes, for effective skip 0.53, serve p95 121\,ms, and success 0.74. At $\gamma{=}0.995$, the trigger fires on 90.4\% of encodes and recovers 1.00 success at approximately twice the dense latency. Even with the clean-signal cost excluded, oracle-driven triggering reaches 0.66--0.92 across the tested thresholds, below the 0.98 from unconditional use of the clean products. These conclusions are limited to the tested triggers on OFT. On CogACT, an oracle trigger at $\gamma{=}0.90$ reaches the dense success rate.

\subsection{Latency and Scheduling}
\label{sec:ledger}

\begin{figure}[ht]
\centering
\includegraphics[width=0.9\columnwidth]{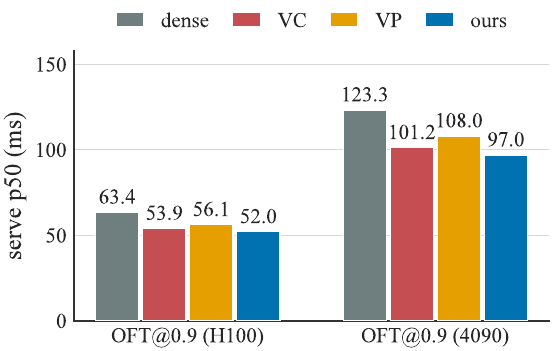}
\caption{Critical-path serve latency on OFT at a 0.9 skip ratio.}
\label{fig:latency}
\end{figure}

\paragraph{Latency} Figure~\ref{fig:latency} compares the 4-suite mean of per-suite serve p50s for the OFT configurations at a 0.9 skip ratio. On H100, \approach{} serves in 52.0\,ms, compared with 63.4\,ms for dense, 53.9\,ms for VC, and 56.1\,ms for VP. On the RTX 4090, the corresponding values are 97.0, 123.3, 101.2, and 108.0\,ms. All comparisons are within the same hardware and harness. The off-path dense pass has a 4-suite mean p50 of 63.6\,ms on H100 and 122.2\,ms on the RTX 4090, both within the ${\sim}400$\,ms actuation window. Per-suite and deadline measurements are reported in the technical appendix. On CogACT pick-coke at a 0.7 skip ratio on H100, dense, VP, and \approach{} take 102.1, 101.2, and 100.0\,ms, respectively, while VC is not applicable because the CogACT port lacks a compute-skipping KV-reuse implementation. Token skipping therefore provides no meaningful serve-time gain on this substrate because its vision encoder and diffusion-transformer action head dominate.

\paragraph{Scheduling} Inline anchoring places its dense passes on the critical path. In the H100 LIBERO-Object timing probe, anchoring every two chunks produces a bimodal profile, with p50 latencies of 64\,ms on anchor steps and 49.7\,ms on sparse steps. In contrast, \approach{} serves every chunk at a p50 of 50\,ms while the refresh runs within the actuation window. Across 561 H100 off-path refreshes, the p99 latency is 79.9\,ms, with no deadline misses.

\subsection{Real-Robot Validation}
\label{sec:real-robot}

We validate serve latency using the same OpenVLA-OFT policy on an AgileX Piper arm with wrist-mounted and global Intel RealSense D455 cameras. Inference runs on an RTX 5880 Ada over a local network. At 10\,Hz, each 8-action chunk executes in 800\,ms, providing a wider refresh window than the ${\sim}400$\,ms in Sec.~\ref{sec:method}. Across 3 tasks with 40 episodes each, \approach{} has the lowest median serve latency, 98.1\,ms versus 125.6\,ms for dense (Table~\ref{tab:realrobot}). The ordering, \approach, VC, VP, then dense, matches Sec.~\ref{sec:ledger}.

\begin{table}[ht]
\centering
\small
\begin{tabular}{@{}lcc@{}}
\toprule
\textbf{method} & \textbf{serve latency (ms)} & \textbf{vs.\ dense} \\
\midrule
dense & 125.6 & -- \\
VC & 102.8 & $\downarrow 18.2\%$ \\
VP & 109.7 & $\downarrow 12.7\%$ \\
\textbf{ours} & \textbf{98.1} & \textbf{$\downarrow 21.9\%$} \\
\bottomrule
\end{tabular}
\caption{Real-robot serve latency on an RTX 5880 Ada. All comparisons are within this platform.}
\label{tab:realrobot}
\end{table}

\subsection{Limitations}
\label{sec:limitations}

\Approach{} is not without cost. It reduces critical-path latency but increases computation to ${\sim}1.6{\times}$ the dense FLOPs and energy by ${\sim}18\%$ per chunk. In addition, available compute and experimental environments limit our reliability evaluation to two policies, 4 LIBERO suites, and 4 SIMPLER tasks, while the physical platform validates latency only. Finally, refresh addresses failures caused by self-harvested gates but not limitations of the gate design or model substrate, and its operating boundary may vary across tasks, substrates, and gate families.

\section{Conclusion}
\label{sec:conclusion}

We identify gate provenance as a central determinant of reliable token skipping in closed-loop VLAs. Gates harvested from served sparse forwards can hide changes in skipped tokens and reinforce their own blind spots under both reuse and deletion. \Approach{} breaks this feedback loop by running a dense pass during action execution, providing the next control step with a clean gate signal and, under reuse, a fresh KV base without extending the critical path. This view shifts the design question from how tokens are skipped to how trustworthy selection signals are maintained. The gate, not the cache.

\bibliography{references}

\appendix
\renewcommand{\thetable}{A\arabic{table}}
\renewcommand{\thefigure}{A\arabic{figure}}
\section{Technical Appendix}
\label{sec:appendix}

\subsection{Baseline Fidelity Audit}
\label{app:fidelity}

\paragraph{VLA-Cache} Our implementation follows VLA-Cache with several disclosed deviations from the released code. We use fixed, controlled skip ratios rather than adaptive reuse counts. We apply full-depth reuse under one mask rather than progressive per-layer schedules that recompute layers 0--1 at each step. The shallow-recompute control in Table~\ref{tab:clean_signal} of the main paper restores this behavior. We combine the pixel and attention terms by summing their ranks rather than thresholding followed by hard exclusion. We aggregate attention over all layers and non-visual queries rather than using text queries from a single layer. We restrict reuse to the agent view, a protocol-wide constraint shared by all arms. All of these choices are held fixed within each paired comparison. An update-ordering slip in the released OFT variant causes the pixel term to compare each frame with itself. We instead implement the intended semantics described in the paper.

\paragraph{VLA-Pruner} Token scores use attention from layer 3, aggregated over all pre-action query rows, and pruning begins at layer 4. The action prior is an EMA($w{=}3,\gamma{=}0.8$) of action-to-vision attention from layer 15 over the true action rows. Scores for dropped patches are zero-scattered exactly as in the released history update. We take the union of the top-$\tilde M$ tokens and apply MMDP with a maximal nearest-neighbor seed. Two sanity checks pass. At the identity budget, the forward matches dense bit-for-bit. Repeated calls with pruning enabled are deterministic. SIMPLER resets are not bit-deterministic across processes, so we run the CogACT equivalence checks within each process. The paired protocol is unaffected. The only retained difference from the released protocol is agent-view-only pruning at a matched ratio, a constraint shared by every arm in the paper. We also disclose one implementation-level confound. The OFT action interface renormalizes proprioception in place, so the off-path refresh pass consumes a renormalized state within each chunk. This confound disfavors the refresh arms and makes their measured gains conservative. The two reuse cells of the controlled design share the confound symmetrically.

\paragraph{SAFE-Pruner} No official implementation was available. We use the published OFT threshold $\gamma{=}0.92$. Full-token key steps compute all-layer pruning saliency, and the forecast linearly extrapolates the two most recent key steps. The trigger $\kappa_t$ compares this forecast with shallow-layer saliency over surviving tokens. We use a same-step variant that favors SAFE-Pruner, so a fired refresh supplies the current action rather than only the next step.

\subsection{Crash Attrition and Sensitivity}
\label{app:crash}

The LIBERO simulator aborts deterministically on specific combinations of episodes and experimental arms. These render-layer failures depend on the arm and concentrate in sparse arms, so we treat the missingness as informative rather than random. Running each episode in a separate process recovers most affected cases. A small residual remains, with Object shallow-recompute completing 99 of 100 episodes. Attrition concentrates in the sparse arms for Long and Goal and in the Spatial VLA-Pruner subset. The Goal VLA-Cache cells have $n{=}57$--$58$, while the Goal VLA-Pruner cells have $n{=}66$. The Spatial VLA-Pruner subset has $n{=}50$. For each head-to-head comparison, both arms use the same set of completed episodes. This preserves pairing within each crash-survivor subset. Absolute success rates are reported with the corresponding $n$ and excluded from the primary conclusions. Under a pessimistic sensitivity analysis that scores every missing episode as a failure of the sparse arm, all 39 collapse verdicts remain unchanged. In contrast, 84 non-degraded cells would flip under this treatment and therefore rely on the complete-case analysis.

\subsection{Latency Details}
\label{app:latency}

\begin{table}[ht]
\centering
\small
\setlength{\tabcolsep}{4pt}
\begin{tabular}{@{}llcccc@{}}
\toprule
\textbf{Hardware} & \textbf{Task} & \textbf{Dense} & \textbf{VC} & \textbf{VP} & \textbf{Ours} \\
\midrule
\multirow{4}{*}{H100} & Object & 61.9 & 52.7 & 54.6 & \textbf{49.0} \\
 & Goal & 63.2 & 53.6 & 56.1 & \textbf{53.3} \\
 & Spatial & 64.0 & 54.6 & 57.4 & \textbf{53.0} \\
 & Long & 64.4 & 54.8 & 56.4 & \textbf{52.5} \\
\midrule
\multirow{4}{*}{RTX 4090} & Object & 129.8 & 107.2 & 112.8 & \textbf{98.2} \\
 & Goal & 123.8 & 105.8 & 111.8 & \textbf{103.2} \\
 & Spatial & 117.9 & 94.8 & 102.3 & \textbf{92.3} \\
 & Long & 121.9 & 97.2 & 105.2 & \textbf{94.2} \\
\bottomrule
\end{tabular}
\caption{Per-task serve latency on LIBERO, measured as in-rollout p50 in milliseconds. All methods use the same harness within each hardware setting. Best results are bold.}
\label{tab:task_invariance}
\end{table}

Latency is CUDA-synchronized wall-clock time from the in-rollout harness after discarding the first two encodes of the first episode for each arm and the first encode of every later episode. Table~\ref{tab:task_invariance} reports task-level p50 separately for H100 and RTX 4090 and compares methods only within the same hardware setting. Our method has the lowest p50 for every task on both GPUs. All other latency and energy measurements use H100. Success-rate rollouts execute refresh synchronously because the simulator imposes no real-time deadline. Main-text VLA-Cache latency uses the compact deployment implementation, whereas the bit-exact injection instrument overwrites KV entries without skipping computation and is used only for equivalence checks. Across 561 off-path refreshes, latency is 63.6, 68.5, 79.9, and 161.3\,ms at p50, p95, p99, and maximum, respectively. All samples finish within the 400\,ms actuation window, supporting the off-path schedule. The instrument arm uses approximately $1.6{\times}$ the LLM FLOPs of dense and consumes 75.5 versus 63.9\,J/chunk. The same accounting applies to the +refresh arms because they use the same dense refresh pass.

\subsection{Qualitative Real-Robot Rollout}
\label{app:realrobot-case}

The platform uses an AgileX Piper 6-DoF arm with a 1.5\,kg payload, 620\,mm reach, and $\pm 0.1$\,mm repeatability. Two Intel RealSense D455 RGB-D cameras provide wrist-mounted and global views at 1920$\times$1080 RGB and 1280$\times$720 depth.

\begin{figure*}[p]
\centering
\includegraphics[width=0.95\textwidth]{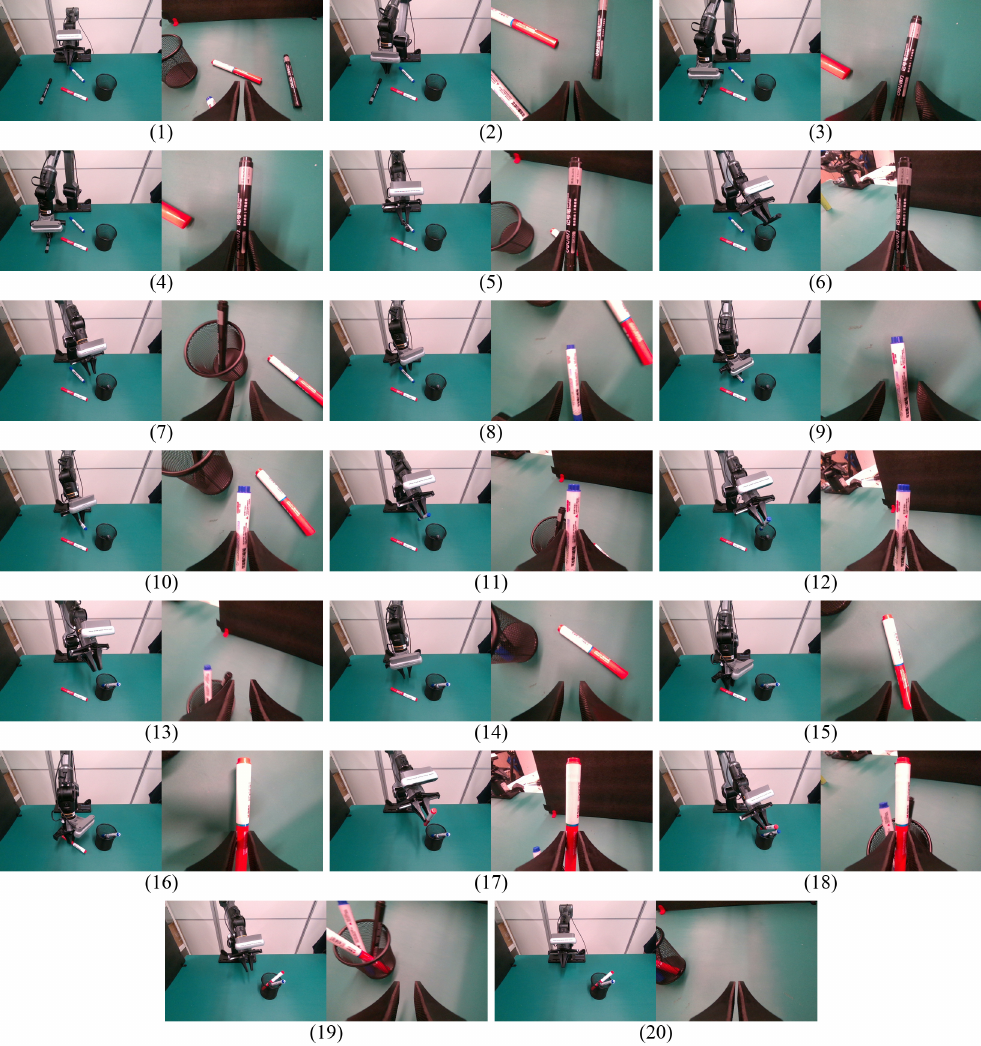}
\caption{Qualitative rollout for ``Put the pens on the table into the pen holder.'' Frames proceed left-to-right and top-to-bottom. Within each frame, the left and right panels show the head- and wrist-camera views, respectively.}
\label{fig:realrobot-case}
\end{figure*}

Figure~\ref{fig:realrobot-case} shows one complete successful physical rollout from the platform of Sec.~\ref{sec:real-robot} of the main paper, sampled at 20 uniformly spaced frames. The task instruction is ``Put the pens on the table into the pen holder.'' The robot places the black, blue, and red pens into the holder in that order. This case illustrates the dual-view observations and closed-loop task execution. It is not used as success-rate evidence.

\subsection{SAFE-Pruner Trigger Sensitivity}
\label{app:safe}

Across 1{,}530 pruned steps in 45 Object failure episodes, the self-harvested consistency $\kappa_t$ exceeds the clean consistency on 67\% of steps, with a median difference of ${+}0.005$. For $\gamma\in\{0.92,0.95,0.97,0.98,0.99\}$, the self-harvested trigger fires on $0.0\%$, $0.5\%$, $5.7\%$, $17.1\%$, and $85.2\%$ of these steps, respectively. The corresponding oracle-trigger rates are $0.1\%$, $7.5\%$, $33.8\%$, $56.5\%$, and $91.2\%$. The closed-loop $\gamma$-sweep and the oracle-driven arm are reported in Sec.~\ref{sec:safe} of the main paper. On CogACT, an oracle trigger at $\gamma{=}0.90$ matches the dense success rate, so conclusions about trigger failure are substrate-specific.

\subsection{Reactive Detector Protocol}
\label{app:governor}

We instrumented the action-level detectors from the silent-failure analysis in closed-loop rollouts that reproduced the collapse. Dense achieved a 1.00 success rate, whereas gated reuse achieved 0.66 and 0.50 at skip ratios of 0.85 and 0.9, respectively. We evaluated two signals. For mid-chunk re-query disagreement, we queried the policy again halfway through chunk execution and measured the $\ell_2$ distance between the two action chunks. Proprioceptive tracking deviation measures the gap between the commanded and reached end-effector poses and is calibrated for each task against dense rollouts. Episode outcomes serve as labels. The early-window score aggregates each signal over the first third of an episode, when intervention remains possible. Across tasks, both signals yielded early-window AUROC values of 0.45--0.65. The full-episode mean of re-query disagreement reached an AUROC of 0.80, but failures ran for all 35 chunks whereas successes ended after a median of 17--19 chunks. After conditioning on episode length, the AUROC fell to 0.26--0.44, showing that the apparent signal was a length artifact. These results show that the measured action-level detector family provides no useful early warning. Token-level triggers are analyzed separately in Sec.~\ref{app:safe}. Feature-probe failure detectors discussed in the main paper target distribution shifts and remain untested for acceleration-induced corruption.

\subsection{Instrument Details}
\label{app:instrument}

We use previous-chunk action attention as the instrument gate and evaluate this empirical choice against alternative gate signals. On Long at a 0.9 skip ratio, clean reuse with the instrument gate reaches 0.88, compared with 0.73 for VLA-Cache with refresh and 0.92 for dense. This comparison suggests that gate design contributes to the remaining gap. When evaluated against action-level ground truth at matched ratios, pixel difference, token drift, and kinematic sweep underperform the instrument gate at every tested point, most by factors of $2$--$7$. In closed-loop Spatial evaluation at a 0.9 skip ratio, optical-flow gating reaches a success rate of 0.74, compared with 0.94 for the instrument gate. Reused-token drift, the proxy underlying pixel and optical-flow gating, does not align with action-level relevance. A patch can change substantially without affecting the action, whereas an apparently stable patch can remain action-critical. Additional results for the instrument arm include the full ratio$\times$provenance cliff grid and the partial-freshness arm. On Object, the partial-freshness arm reaches success rates of 0.98, 0.92, and 0.52 at skip ratios of 0.7, 0.9, and 0.97, respectively, interpolating between self-harvested and clean gates. The remaining controls include inline anchoring (Sec.~\ref{sec:ledger} of the main paper) and a one-step observation-delay probe that reduces the Spatial success rate by 0.08 despite dense computation.

\end{document}